\documentclass[11pt]{article}

\usepackage[final]{acl}

\usepackage{times}
\usepackage{latexsym}

\usepackage[T1]{fontenc}

\usepackage[utf8]{inputenc}

\usepackage{microtype}

\usepackage{inconsolata}

\usepackage{graphicx}

\usepackage{algorithmic}
\usepackage{algorithm}

\usepackage{amsmath}
\usepackage{amssymb}   

\usepackage{booktabs}
\usepackage{enumitem}

\usepackage{tabularx}
\usepackage[normalem]{ulem}

\usepackage[most]{tcolorbox}

\definecolor{usercolor}{RGB}{235, 242, 250}
\definecolor{agentcolor}{RGB}{255, 255, 255}
\definecolor{toolcallcolor}{RGB}{255, 250, 240}
\definecolor{tooloutputcolor}{RGB}{245, 245, 245}
\definecolor{memorycolor}{RGB}{248, 242, 255}
\definecolor{finalanswercolor}{RGB}{240, 255, 245}

\newtcolorbox{dialoguebox}[3][]{
    breakable, 
    enhanced,
    before skip=3pt, 
    after skip=3pt,
    left=3pt, right=3pt, top=3pt, bottom=3pt,
    colframe=#2!60!black, 
    colback=#2,
    title={\textbf{#3}}, 
    fonttitle=\sffamily\tiny,
    fontupper=\footnotesize, 
    arc=2pt,       
    boxrule=0.5pt,
    parbox=false, 
    #1
}

\author{
  \textbf{Yuxiang Zhang\textsuperscript{1}},
  \textbf{Jiangming Shu\textsuperscript{1}},
  \textbf{Ye Ma\textsuperscript{2}},
  \textbf{Xueyuan Lin\textsuperscript{3,2}},
  \textbf{Shangxi Wu\textsuperscript{4}},
  \textbf{Jitao Sang\textsuperscript{1,5}\thanks{Corresponding author.} }
\\
  \textsuperscript{1}School of Computer Science and Technology, Beijing Jiaotong University \\
  \textsuperscript{2}Hithink Research \\
  \textsuperscript{3}The Hong Kong University of Science and Technology (Guangzhou) \\  
  \textsuperscript{4}Huawei Noah's Ark Lab  
  \textsuperscript{5}Peng Cheng Lab \\
\\
  \texttt{\{yuxiangzhang, jiangmingshu, jtsang\}@bjtu.edu.cn} \\
  \texttt{maye@myhexin.com, linxy59@mail2.sysu.edu.cn, wushangxi1@huawei.com}
}

\title{Memory as Action: Autonomous Context Curation for Long-Horizon Agentic Tasks}
\usepackage{multirow}
\begin{document}
\maketitle
\begin{abstract}
Long-context Large Language Models, despite their expanded capacity, require careful working memory management to mitigate attention dilution during long-horizon tasks. Yet existing approaches rely on external mechanisms that lack awareness of the agent's reasoning state, leading to suboptimal decisions. We propose \uline{Mem}ory-as-\uline{Act}ion (MemAct), a framework that treats working memory management as learnable policy actions. By formulating context management as in-place editing operations (deletion, insertion), MemAct enables joint optimization of information retention and task performance through end-to-end reinforcement learning. To address the computational challenges of dynamic context updates, we introduce Dynamic Context Policy Optimization, which restores training efficiency without compromising reasoning integrity. Experiments show that MemAct-RL-14B matches the accuracy of models $16\times$ larger while reducing average context length by 51\%, with learned strategies that adapt to model capabilities and generalize across task complexities.
The code and datasets are available at \url{https://github.com/ADaM-BJTU/MemAct}.
\end{abstract}

\section{Introduction}
For agentic tasks demanding long-horizon reasoning and complex tool use, such as deep research and software engineering agents~\citep{wei2025browsecomp,jimenez2024swe}, the effectiveness of a Large Language Model (LLM) is fundamentally constrained by what information resides in its context. The agent's working memory is realized as the input context, a sequence of tokens encoding the interaction history available at each decision step. However, left unmanaged, this context inevitably saturates with irrelevant information, triggering attention dilution that buries critical signals and results in ``lost-in-the-middle'' behavior~\citep{liu-etal-2024-lost}. The critical bottleneck thus shifts from merely expanding memory capacity to actively curating its contents. We term this challenge \textit{Context Curation}: the process of strategically selecting, integrating, and pruning information to maintain a focused and goal-relevant reasoning trace. 

\begin{figure}[t]
  \centering
  \includegraphics[width=0.86\linewidth]{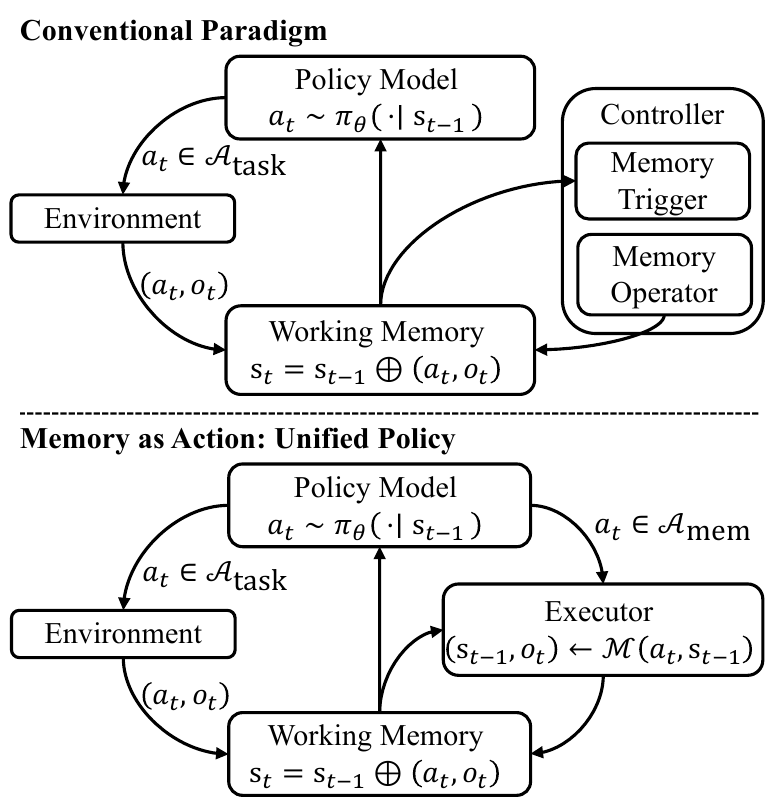}
\caption{\textbf{Comparison of context management paradigms.} 
\textit{Top:} Conventional approaches decouple memory management from the policy, where an external controller with heuristic triggers and operators governs context independently. 
\textit{Bottom:} MemAct unifies task actions $\mathcal{A}_{\text{task}}$ and memory actions $\mathcal{A}_{\text{mem}}$ within a single policy $\pi_\theta$, enabling end-to-end optimization.}

  \label{fig:memact-overview}
\end{figure}

Recent advances in long-context methods have successfully expanded the capacity of an agent's working memory~\citep{peng2023yarn, deepseekai2024deepseekv32}.  
However, simply increasing the context window does not guarantee improved reasoning performance. The effectiveness of long-context models is fundamentally determined by Context Engineering~\citep{mei2025survey}, which refers to the deliberate curation and structuring of information to ensure the most relevant evidence is accessible at the right time.  
The dominant approach to context engineering today relies on a workflow of heuristic rules~\citep{packer2023memgpt, xu2025mem, 2025mem1learningsynergizememory}. These designs decouple memory management from the agent's core reasoning policy, preventing the end-to-end optimization of information retention against task performance. 

We bridge this divide by reconceptualizing context management as an intrinsic, learnable primitive rather than a policy-agnostic mechanism. This shift is non-trivial, as it requires agents to navigate the inherent trade-offs between task performance and context efficiency through joint optimization. We propose \uline{Mem}ory-as-\uline{Act}ion (MemAct), a framework that treats context curation as a set of learnable actions within a unified policy space. Rather than passively accumulating an ever-growing prefix, the agent learns to decide when to retain, compress, or discard segments of history, or synthesize content to maintain context coherence. These transformations are applied through explicit function-call actions, enabling the agent to develop memory strategies that improve reasoning efficiency, as shown in Fig.~\ref{fig:memact-overview} for a schematic overview.

To learn such memory-editing actions for dynamic control, we adopt an end-to-end reinforcement learning approach. However, this flexibility introduces a critical training challenge. Causal LMs assume monotonic context growth, computing states over preceding sequences. When MemAct updates context, this assumption breaks: deleted content already influenced subsequent token representations, creating a train-inference mismatch requiring physical trajectory restructuring.

For the sake of reconciling dynamic memory with large-scale training efficiency, we also propose \textit{Dynamic Context Policy Optimization} (DCPO). DCPO restores training feasibility by logically segmenting fractured trajectories, enabling the policy to be optimized end-to-end within standard, highly optimized infrastructure without bespoke modifications.
In summary, our core contributions are:

\begin{itemize}[leftmargin=*, noitemsep, topsep=2pt]
    \item \textbf{Paradigm:} We propose the Memory-as-Action paradigm, which shifts working memory management from external mechanisms or fixed routines to an intrinsic, learnable policy capability. By integrating memory editing as actions within a unified policy space, MemAct enables agents to autonomously balance context curation and task execution through end-to-end optimization.

    \item \textbf{Method:} We contribute two technical components: (1) a Markov Decision Process (MDP) formulation with ID-based addressable decision sequences and the \texttt{Prune\&Write} operator, enabling precise, fine-grained working memory editing; (2) DCPO, a trajectory segmentation algorithm that reconciles dynamic context updates with efficient RL training on standard RL infrastructure.
    \item \textbf{Empirical Validation:} We demonstrate that learned memory strategies exhibit efficiency, adaptivity, and generalizability: MemAct-RL-14B matches Qwen3-235B accuracy using 49\% of the average context length, distinct strategies emerge tailored to different backbone models, and learned policies transfer across task complexity and domains. These findings establish autonomous context management as a formidable, scalable, and model-intrinsic capability.
\end{itemize}

\section{Related Work}
Effective long-horizon reasoning demands active management of working memory, which serves as the evolving workspace that maintains task-relevant context~\citep{hu2025memory, hu2025hiagent}. Existing approaches bifurcate into two paradigms. One line of work treats context as a constrained resource, applying token-level compression~\citep{jiang2024longllmlingua, zhang2023ho}, selective pruning~\citep{li2023compressing}, or periodic summarization~\citep{lu2025scaling, wu2025resum} to fit information within fixed windows. While computationally efficient, these methods operate without awareness of the agent's reasoning state, risking the loss of semantically critical dependencies. An alternative paradigm delegates memory operations to external controllers~\citep{packer2023memgpt, xu2025mem, chhikara2025mem0}, which manage structured formation, evolution, and retrieval. However, this decoupled architecture prevents joint optimization of information retention and downstream task performance.
Recent efforts explore RL to internalize memory as a learnable capability~\citep{yan2025memory, yu2025memagent, 2025mem1learningsynergizememory}. Yet these approaches typically impose rigid constraints: mandatory per-step compression or coarse-grained retrieval that treats context as a monolithic buffer. In contrast, MemAct formulates working memory management as fine-grained, addressable editing actions within a unified policy, enabling the agent to perform selective, surgical action aligned with its evolving reasoning needs.

\section{Method}
\label{sec:method}
\begin{figure*}[t]
  \centering
  \includegraphics[width=0.88\textwidth]{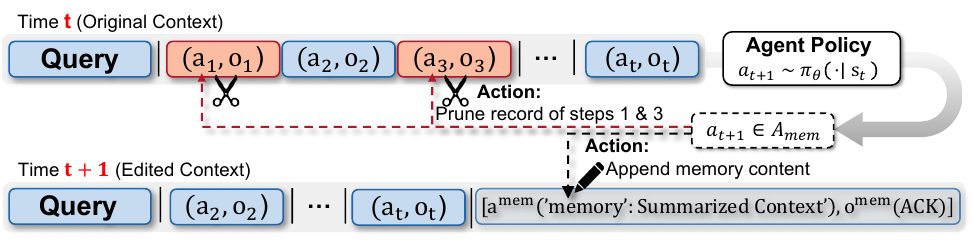}
\caption{
  \textbf{Workflow of MemAct: Autonomous Context Management.} 
  At timestep $t$, the policy generates a \texttt{Prune\&Write} action that specifies (1) which historical turns to remove (indices 1 and 3), and (2) a synthesized memory note containing summaries or key facts.
  The action itself is retained \textbf{in-place} as persistent memory, transforming the original context into a compact state $s_{t+1}$ for subsequent reasoning.
}
  \label{fig:inference_process}
\end{figure*}

This section presents the MemAct framework in three parts: 
an operational overview of autonomous context management (\S\ref{ssec:overview}), 
MDP formalization of the problem (\S\ref{ssec:mdp_formalization}), 
and Dynamic Context Policy Optimization (DCPO), our training algorithm for non-sequential context updates (\S\ref{ssec:dcpo}).

\subsection{Overview}
\label{ssec:overview}
MemAct internalizes context management by integrating it directly into the policy's action space. Unlike external memory systems that operate via fixed heuristics, MemAct enables the policy to autonomously learn when and how to curate its working memory using the \texttt{Prune\&Write} operator, a unified primitive for in-place context editing.

As illustrated in Fig.~\ref{fig:inference_process}, rather than passively accumulating all interaction history, the agent intersperses memory actions to enact context updates. The workflow consists of three key steps:
\begin{enumerate}[leftmargin=*, noitemsep, topsep=2pt]
    \item \textbf{Action Selection:} Given the current state $s_t$, the agent samples an action $a_t \sim \pi_\theta(a | s_t)$ from the augmented action space $\mathcal{A} = \mathcal{A}_{\text{task}} \cup \mathcal{A}_{\text{mem}}$. This selection is implicit in the model's generation, allowing it to dynamically switch between reasoning and context management.
    \item \textbf{Operation Parameterization:} If a memory action is selected, the model instantiates the operation by generating its specific parameters: a set of index IDs to prune and a text field of memory content to summarize or reflect.
    
    \item \textbf{Execution:} The system executes the pruning based on the target IDs. Crucially, the $a^{\text{mem}}$ action record (containing the new memory content) is appended in-place, making the memory contents mutable for future update.
\end{enumerate}

This process recursively transforms working memory into a curated state, continuously keeping critical information within a bounded context.

\subsection{MDP Formulation}
\label{ssec:mdp_formalization}

We model the agent's interaction as a Markov Decision Process, formalizing working memory as a sequence of uniquely addressable interaction turns.

\begin{itemize}[leftmargin=*, noitemsep, topsep=2pt]
    \item \textbf{State}: The state $s_t$ is the current working memory $H_t$, represented as a sequence of interaction records $H_t = [z_1, z_2, \dots, z_k]$. Each record $z_i = (a_i, o_i, \text{id}_i)$ comprises an action, its  observation, and a unique identifier $\text{id}_i$ that ensures precise addressing regardless of context shifts.
    
    \item \textbf{Action Space}: $\mathcal{A} = \mathcal{A}_{\text{task}} \cup \mathcal{A}_{\text{mem}}$. 
    \begin{itemize}[leftmargin=*, noitemsep, topsep=2pt]
        \item $\mathcal{A}_{\text{task}}$: Standard environment interactions (e.g., search, web browser).
        \item $\mathcal{A}_{\text{mem}}$: The context management operator. A memory action takes the form $a^{\text{mem}} = (\mathcal{I}_{\text{target}}, c)$, where $\mathcal{I}_{\text{target}}$ is the set of IDs to remove and $c$ represents the generated memory content. This content synthesizes summarization, reflection, and planning to ensure reasoning continuity despite memory pruning.
    \end{itemize}
    
    \item \textbf{Transition}: The transition dynamics depend on the action type:
    \begin{itemize}[leftmargin=*, noitemsep, topsep=2pt]
        \item \textit{Task Action}: $H_{t+1} = H_t \oplus (a_t, o_t, \text{id}_{\text{new}})$.
        \item \textit{Memory Action}: Given $a_t = (\mathcal{I}_{\text{target}}, \text{content})$, the system executes an ID-based filtering operation:
        \begin{equation}
        \label{eq:trans}
        \begin{split}
            H_{t+1} &= \{z_i \in H_t \mid \text{id}_i \notin \mathcal{I}_{\text{target}}\} \\
                    &\quad \oplus (a_t, o_{\text{status}}, \text{id}_{\text{mem}})
        \end{split}
        \end{equation}
        The record of the memory action is appended in-place, ensuring the curated summary remains addressable.
    \end{itemize}
    
    \item \textbf{Objective}: Learn a policy $\pi_\theta(a|H_t)$ that maximizes the expected cumulative reward.
\end{itemize}

\subsection{Dynamic Context Policy Optimization}
\label{ssec:dcpo}
Optimizing MemAct is primarily hindered by a structural misalignment between generated tokens and their corresponding generative contexts. While conventional policy gradient objectives presuppose a strictly monotonic, incremental history to maximize computational efficiency, the \texttt{Prune\&Write} operator introduces non-continuous trajectories where $H_{t+1} \not\supseteq H_t$ (see Fig.~\ref{fig:inference_process}). This departure from standard auto-regressive assumptions causes naive backpropagation to compute gradients against historically mismatched states, inevitably resulting in severely biased credit assignment. To resolve this, DCPO restructures these non-continuous trajectories into a series of logically consistent, independent segments, thereby restoring the intrinsic causal structure required for stable optimization.

\paragraph{Why Not Simple Attention Masking?}
While attention masking offers a seemingly straightforward approach to managing memory deletions, it is fundamentally incompatible with the causal nature of LLMs. In the architecture of causal language models, the latent representation of each token encapsulates information from its entire antecedent sequence. Thus, the influence of a semantically ``deleted'' token is already encoded into the key-value states of all subsequent tokens generated prior to the deletion. This creates an irreconcilable mismatch: the model's internal states remain \textbf{conditioned on information that is semantically absent but physically persistent} in the historical KV cache. To truly learn from the post-edit history, the trajectory must be physically reconstructed to sever these causal dependencies. Furthermore, production-grade inference engines are architecturally tailored for monotonic context expansion, making non-linear cache modifications or frequent recomputations computationally prohibitive.

\subsubsection{Trajectory Segmentation}
\label{sssec:segmentation}

To resolve context misalignment while preserving training scalability, DCPO logically partitions the trajectory at each memory edit point. Let $t^{\text{mem}}_1, \ldots, t^{\text{mem}}_K$ denote the timesteps of memory actions, with $t^{\text{mem}}_0 = 0$ and $t^{\text{mem}}_{K+1} = T$. The trajectory is re-organized into $K+1$ independent segments $\{\sigma_i\}_{i=0}^K$. For clarity, we represent each segment as a tuple:
\begin{equation}
    \sigma_i = (C_i, \mathbf{y}_i)
\end{equation}
where $C_i = H_{t^{\text{mem}}_i}$ is the fixed context prefix at the start of the segment, and $\mathbf{y}_i = \mathbf{y}_{t^{\text{mem}}_i+1:t^{\text{mem}}_{i+1}}$ is the subsequent token sequence. \textbf{The crucial insight is that within any segment $\sigma_i$, the context prefix $C_i$ remains fixed}, ensuring that the sequential dependency holds locally. During training, we generate $N_{\text{traj}}$ full trajectories for each prompt and sample a subset of segments $\Sigma(\tau) \subseteq \{\sigma_i\}$ for optimization using a trajectory-based round-robin strategy to ensure balanced coverage.

\subsubsection{Reward Design}
\label{sssec:reward}

Each full trajectory $\tau$ is assigned a sparse, terminal reward $R(\tau)$ contingent on its final outcome:
\begin{equation}
R(\tau) =
\begin{cases}
    r_{\text{task}} & \text{if the task succeeds,} \\
    r_{\text{pen}} & \text{if a constraint is violated,} \\
    0 & \text{otherwise.}
\end{cases}
\end{equation}
Here, $r_{\text{task}} > 0$ denotes the incentive for successful completion, while $r_{\text{pen}} < 0$ penalizes constraint violations (e.g., exceeding the maximum context length). This sparse signal encourages the policy to jointly optimize for both functional correctness and resource efficiency.

\subsubsection{Reward Attribution and Optimization}
\label{sssec:objective}

Since the final outcome $R(\tau)$ depends on the collective sequence of memory edits and generations, we adopt a global credit assignment strategy where each sampled segment $\sigma \in \Sigma(\tau)$ inherits the trajectory-level advantage $A(\tau)$. This advantage is computed using the group-relative normalization scheme:
\begin{equation}
    A(\tau) = \frac{R(\tau) - \text{mean}(\mathcal{R}_u)}{\text{std}(\mathcal{R}_u) + \epsilon}
\end{equation}
where $\mathcal{R}_u$ is the set of rewards for all $N_{\text{traj}}$ trajectories sampled for prompt $u$. The policy is optimized by minimizing the following objective:
\begin{align}
\mathcal{L}(\theta) &= -\,\mathbb{E}_{u\sim\mathcal{D}}\!\left[ \frac{1}{|\mathcal{G}(u)|} \sum_{\tau\in\mathcal{G}(u)} \mathcal{L}_\tau \right], \\
\mathcal{L}_\tau &= \sum_{(C, \mathbf{y}) \in \Sigma(\tau)} \mathcal{J}_{\text{clip}}(\mathbf{y} \mid C, A(\tau))
\end{align}
where $\Sigma(\tau)$ denotes the set of logically consistent segments reconstructed from trajectory $\tau$, and $\mathcal{J}_{\text{clip}}$ denotes the clipped surrogate objective following the GRPO~\citep{shao2024deepseekmath}. This formulation ensures that gradients are computed against the correctly reconstructed context mapped to each training segment while remaining concise.

\section{Experiments \& Results}
\subsection{Datasets}
\label{ssec:datasets}

We evaluate MemAct using synthetic data and public benchmarks to assess its reasoning efficiency. Our analysis focuses on two pivotal dimensions: maintaining accuracy under context pressure and generalizing from low-complexity training tasks to unseen, high-complexity inference scenarios.

\subsubsection{Evaluation Benchmarks}

\paragraph{Multi-objective Tasks}
To test the agent's long-range reasoning and memory management, we built a multi-objective QA dataset based on HotpotQA, following the construction method in \citep{2025mem1learningsynergizememory}. In each task, the agent must answer several independent sub-questions to provide a single final answer. We evaluate the model on test sets with up to 8 objectives, with 200 samples at each level.

\paragraph{Single-Objective Tasks} To evaluate MemAct's robustness across different reasoning lengths, we selected a diverse range of benchmarks, from standard multi-hop queries to complex, long-horizon reasoning tasks. This set includes 2WikiMultihopQA~\citep{ho2020constructing}, Bamboogle~\citep{press2022measuring}, HotpotQA~\citep{yang2018hotpotqa}, and Musique~\citep{trivedi2022musique}, as well as the more challenging Frames~\citep{krishna2024fact} and BrowseComp-Plus~\citep{chen2025BrowseCompPlus}.

\subsubsection{Training Data Construction}
This section introduces the data composition in the training process, and detailed statistics for SFT and RL can be found in Table~\ref{tab:train_data_stat} in the Appendix. We also explain our data construction and how the training setup is used to test model generalization.

\paragraph{Synthetic Data for SFT Initialization.}
Preliminary experiments showed that even frontier models, such as OpenAI o3 and DeepSeek-V3.1, struggle with managing working memory automatically. Common failures include ignoring the tool entirely, invoking it repetitively, or losing track of the flow after a memory update. To fix this, we use DeepSeek-V3.1 to synthesize training trajectories through a staged prompting method. When the context length is between 8K and 16K tokens, we insert a message suggesting the model check if a memory action is needed. Once the context exceeds 16K tokens, we use strict messages to force the operation. We only keep successful trajectories where the final answer is correct, and we remove the injected hints in the final SFT dataset to ensure the model learns to act independently.

\paragraph{RL Dataset and Complexity Scaling.}
The RL phase combines single-objective tasks from Asearcher~\citep{gao2025turnsunlockinglonghorizonagentic} and synthesized multi-objective tasks. We deliberately limit the training tasks to at most three objectives. This setup allows us to test the model's generalization: by training only on simpler cases, we can verify that the gains on harder tasks (4 to 8 objectives) come from a learned general working memory management strategy rather than memorizing training patterns.

\subsection{Evaluation Metrics} \label{ssec:metrics}
We measure Task Accuracy using an LLM-based evaluator~\citep{openai2025gptoss120bgptoss20bmodel} with a three-pass consensus protocol. If any of the three checks fails, the answer is marked as incorrect. For single-objective benchmarks, this metric is the success rate; for multi-objective tasks, it is the average success rate across all sub-objectives. We also track the Solved Sub-objective Count to evaluate reasoning depth. To measure efficiency, we record the total number of tokens used and the frequency of tool calls.

\subsection{Baselines}
\label{ssec:baselines}

We compare MemAct against three types of baselines, ranging from models using full context to those with externally-managed or RL-based agents. Unless otherwise specified, all baselines are implemented using their default configurations and the same LLM as MemAct to ensure a fair comparison.

\paragraph{Full-Context Baseline} 
We use Qwen3-235B-A22B-Instruct as a full-context baseline. With no memory pruning, it represents the performance upper bound for our evaluation.

\paragraph{Externally-managed Strategies}
These methods manage memory using fixed rules or external systems, rather than the agent's own policy:
\begin{itemize}[leftmargin=*, noitemsep, topsep=2pt]
    \item \textbf{Sliding Window}: This method naively keeps only the most recent 8K tokens and discards older context once the limit is reached.
    \item \textbf{Summarization}: This method adds a short summary of the discarded content, generated by the model itself, to the Sliding Window approach.
    \item \textbf{A-MEM}~\citep{xu2025mem}: A system that organizes historical experiences into interconnected networks through dynamic linking and allows memories to evolve as new information arrives.
\end{itemize}

\paragraph{Learning-based Agents}
We also compare MemAct against other agents that learn to manage context through training:
\begin{itemize}[leftmargin=*, noitemsep, topsep=2pt]
    \item \textbf{MEM1}~\citep{2025mem1learningsynergizememory}: An RL-based baseline that also learns memory actions through training. It follows a fixed schedule where state compression is triggered at every step. We re-trained it using our training data.
    \item \textbf{Tongyi-DeepResearch}~\citep{tongyidr}: A 30B-parameter model specialized in autonomous web research, optimized via reinforcement learning to handle long-horizon tasks.
    \item \textbf{Search-R1}~\citep{jin2025search}: This baseline is essentially MemAct without the memory action capability. It follows the same training pipeline using GRPO on the same dataset, but \textbf{cannot perform memory actions}. 

\end{itemize}

\subsubsection{Implementation Details}
\paragraph{Model and Training.}
We use Qwen2.5-7B-Instruct and Qwen2.5-14B-Instruct~\citep{team2024qwen2} as base models. In the SFT stage, we train the model for 6 epochs with a batch size of 256 and a learning rate of $5 \times 10^{-5}$, using cosine decay and 10\% warm-up. The model obtained after this stage is denoted as MemAct-SFT. In the RL stage, we use the DCPO algorithm with a batch size of 128 and a constant learning rate of $1 \times 10^{-6}$. The final model after reinforcement learning is denoted as MemAct-RL. Both stages are optimized by AdamW. All experiments are conducted on NVIDIA H100 GPUs. Following the strategy in \S\ref{sssec:segmentation}, we set $N_{\text{traj}}=5$ and $N_{\text{seg}}=12$. Tasks are limited to a maximum of 40 steps, including memory actions.

\paragraph{Reward Configuration.}
The numerical specifications for the reward function $R(\tau)$ are as follows. We assign $r_{\text{task}}=+1.0$ for successful task completion and $r_{\text{pen}}=-0.1$ for any violation of operational constraints, such as exceeding the 20K token context limit or the 40-step execution threshold. All other outcomes result in a zero reward. Task success is determined by an LLM-based evaluator that assesses the semantic consistency between the agent's final answer and the ground truth.

\subsection{Main Results}

\label{sec:main_results}
\begin{figure}[t]
  \centering
  \includegraphics[width=\linewidth]{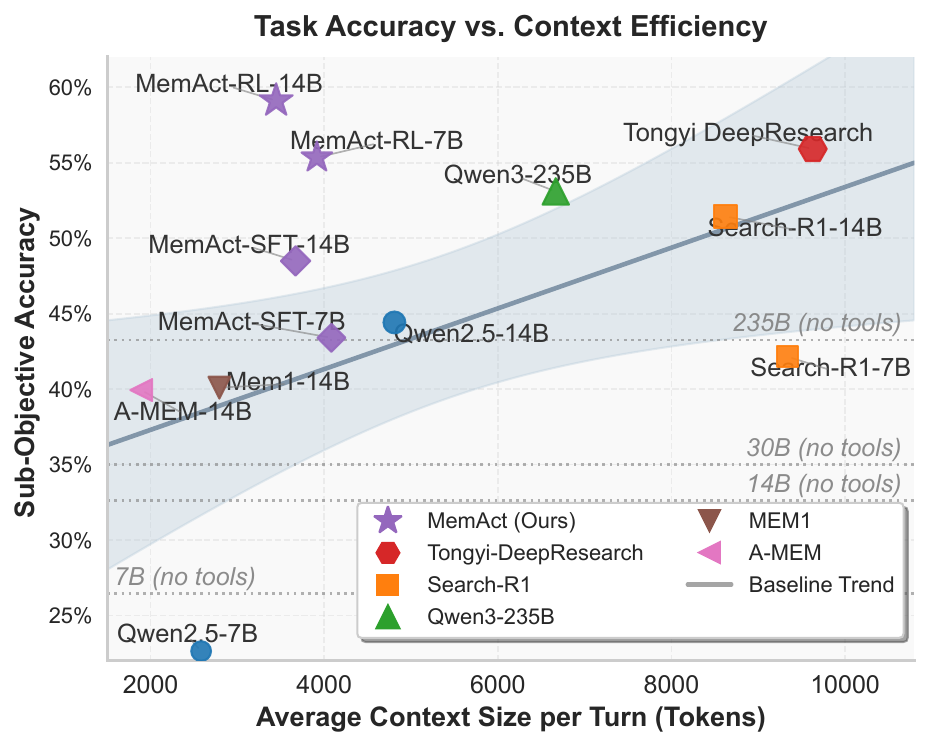}
    \caption{\textbf{Accuracy--Efficiency Trade-off.} 
MemAct variants (stars) occupy the Pareto frontier, achieving competitive accuracy with significantly reduced context size. Dashed lines show no-tool baselines at different model scales for reference.}
\label{fig:multi_object_qa_results}
\end{figure}

\begin{figure}[t!]
  \centering
  \includegraphics[width=\linewidth]{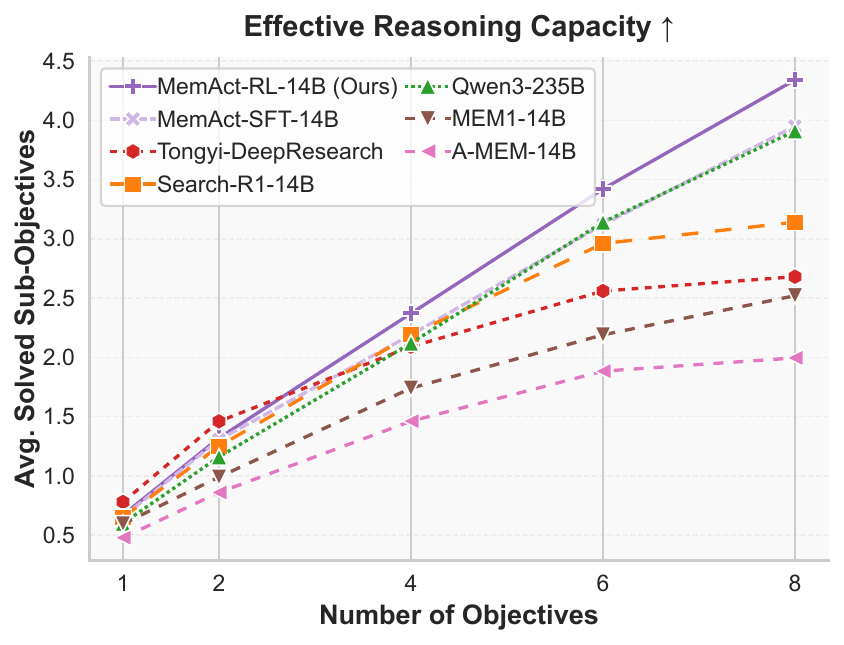} 
    \caption{
      \textbf{Reasoning Stability under Complexity.} 
      As task complexity increases (number of sub-objectives), baselines suffer from performance saturation, while \texttt{MemAct-RL} maintains a more stable trajectory.
    }
\label{fig:reasoning_capacity}
\end{figure}

\begin{table*}[t]
\centering
\small
\setlength{\tabcolsep}{3.0pt}
\renewcommand{\arraystretch}{1.15}

\caption{\textbf{Main results on Single-Objective and Multi-Objective benchmarks.} Except for Qwen3-235B, all methods use Qwen2.5-14B-Instruct. Performance is reported as \textbf{Task Accuracy}, defined as the success rate of single objective tasks and the average success rate of sub-objectives for Multi-Objective tasks. ``Cost'' denotes average token consumption ($\times 10^4$). \textbf{Bold} and \underline{underlined} values indicate the best and second-best performance.}
\label{tab:main_results}
\begin{tabular}{l|cccccc|cccccc}
\toprule
\multirow{2}{*}{\textbf{Method}} 
 & \multicolumn{6}{c|}{\textbf{Single-Objective Tasks} {\scriptsize (Difficulty: Low $\rightarrow$ High)}} 
& \multicolumn{6}{c}{\textbf{Multi-Objective Tasks}} \\
\cmidrule(lr){2-7} \cmidrule(lr){8-13} 
 & \textbf{2Wi.} & \textbf{Hot.} & \textbf{Bam.} & \textbf{Fra.} & \textbf{Bro.} & \textbf{Avg.}
 & \textbf{2-obj.} & \textbf{4-obj.} & \textbf{6-obj.} & \textbf{8-obj.} & \textbf{Avg.} & \textbf{Cost($\times 10^4$)} \\
\midrule

Qwen3-235B
& 0.603 & 0.690 & 0.505 & 0.430 & 0.268 & 0.500
& 0.580 & 0.531 & 0.523 & 0.489 & 0.531 & 16.7 \\

Qwen2.5-14B
& 0.580 & 0.655 & 0.488 & 0.275 & 0.111 & 0.421
& 0.510 & 0.466 & 0.422 & 0.378 & 0.444 & 6.5 \\
\midrule

Sliding Window 
& 0.535 & 0.560 & 0.472 & 0.215 & 0.085 & 0.373
& 0.517 & 0.513 & 0.474 & 0.475 & 0.495 & \textbf{3.8} \\

Summarization 
& 0.540 & 0.692 & 0.442 & 0.335 & 0.120 & 0.426
& 0.540 & 0.498 & 0.495 & 0.471 & 0.501 & 4.1 \\

A-MEM
& 0.528 & 0.591 & 0.453 & 0.250 & 0.094 & 0.383
& 0.430 & 0.385 & 0.364 & 0.339 & 0.399 & \underline{3.9} \\

\midrule

Search-R1 (SFT + RL)
& \textbf{0.775} & \textbf{0.723} & \textbf{0.624} & \underline{0.376} & \underline{0.177} & \underline{0.535}
& 0.625 & 0.546 & 0.493 & 0.393 & 0.514 & 19.3 \\

MEM1 (RL)
& 0.565 & 0.660 & 0.470 & 0.285 & 0.141 & 0.424
& 0.494 & 0.435 & 0.357 & 0.293 & 0.435 & 5.0 \\

MemAct (SFT)
& 0.764 & 0.705 & 0.616 & 0.359 & 0.160 & 0.521
& \underline{0.652} & \underline{0.547} & \underline{0.520} & 0.493 & \underline{0.553} & 9.4 \\

MemAct (SFT+RL)
& \underline{0.767} & \underline{0.710} & \underline{0.618} & \textbf{0.385} & \textbf{0.207} & \textbf{0.537}
& \textbf{0.660} & \textbf{0.591} & \textbf{0.570} & \textbf{0.543} & \textbf{0.591} & 8.2 \\

\quad \textit{w/ Fixed Update}$^\dagger$
& - & - & - & - & - & - 
& 0.672 & 0.581 & 0.557 & 0.516 & 0.582 & 7.0 \\
\bottomrule
\multicolumn{12}{l}{\footnotesize $^\dagger$Enforces a static memory schedule with a fixed update interval of 5 turns.}
\end{tabular}
\end{table*}
As shown in Figure~\ref{fig:multi_object_qa_results} and Table~\ref{tab:main_results}, \textbf{MemAct variants (stars) consistently occupy the top-left Pareto frontier}, which indicates they achieve higher accuracy with much smaller context sizes than all baselines. Specifically, MemAct-RL-14B reaches the highest multi-objective accuracy of 59.1\%, outperforming the much larger Qwen3-235B (53.1\%) and the specialized Tongyi-DeepResearch (56.0\%). Notably, MemAct-RL-14B maintains this lead while operating with a lean average input context length of only 3,500 tokens per step, which is nearly 50\% shorter than Qwen3-235B and 60\% shorter than Search-R1-14B.

This performance lead is accompanied by a significant reduction in total computational cost. Table~\ref{tab:main_results} shows that MemAct-RL-14B uses only $8.2 \times 10^4$ tokens on average. This total cost is about 51\% lower than that of Qwen3-235B ($16.7 \times 10^4$) and 57\% lower than Search-R1-14B ($19.3 \times 10^4$). While some fixed-rule baselines like A-MEM ($3.9 \times 10^4$) have lower costs, their accuracy is much lower at 39.9\%. These smaller context sizes also provide MemAct with a major advantage in inference latency, as analyzed below.

\paragraph{Latency and Efficiency.}
We measured the latency across 2,000 trajectories using the SGLang inference engine~\citep{zheng2024sglang}. Results show that MemAct-RL-7B reduces total duration by 40\% compared to Search-R1, even though Search-R1 performs fewer tool calls. This speedup comes from two main factors. First, by maintaining a compact average context size, MemAct reduces pre-fill time and prevents the decoding speed from slowing down. Since memory updates are sparse, the context history remains stable, which increases the prefix cache hit rate. Second, MemAct eliminates the need for auxiliary inference passes. Unlike methods such as A-MEM that re-process the entire context to generate summaries or evaluate states, MemAct executes memory actions inline within the reasoning flow. This approach avoids the high cost of separate maintenance steps and saves significant time during long-range reasoning.

\subsection{Ablation Analysis} \label{ssec:ablation}
We conduct an ablation analysis to investigate how making memory management an active policy decision, rather than a passive or fixed process, contributes to the overall performance of MemAct.

\paragraph{Effect of Active Context Management.}
Active memory management is essential for maintaining reasoning quality beyond simple token savings. Search-R1 serves as an ablation of MemAct without memory management and retains all information during reasoning. Although reinforcement learning enhances reasoning in Search-R1, the absence of memory actions leads to excessive token growth and context noise. As shown in Table~\ref{tab:tool_usage_profile}, MemAct-RL-7B performs 28.9 total tool calls on average, exceeding the 23.5 calls of Search-R1-7B. Despite this increased activity, MemAct maintains higher accuracy by removing irrelevant history through proactive memory actions.

\paragraph{Comparison of Learning and Fixed Policies.}
Reinforcement learning is indispensable for optimizing memory decisions, as MemAct-RL improves multi-objective accuracy from 0.485 to 0.591 compared to its SFT version. The inherent limitations of rigid schedules are further shown by the \texttt{Fixed-Interval} baseline, which executes memory actions every five turns. Although this rule reduces context size, it causes a performance drop compared to MemAct, especially in complex tasks. As reported in Table~\ref{tab:main_results}, the accuracy of the fixed policy falls behind MemAct on 8-objective tasks. This suggests that static schedules often delete critical information, while MemAct learns to synchronize memory actions with the reasoning process.

\subsection{Scalability and Generalization}
\label{ssec:scalability_analysis}
We evaluate how MemAct performs when tasks become more complex or move to new environments. Specifically, we focus on discovering better management strategies without human intervention.

\paragraph{Scaling to Complex Tasks.}
We tested the models on tasks with an increasing number of sub-objectives. As shown in Fig.~\ref{fig:reasoning_capacity}, baselines reach a performance bottleneck when tasks exceed four objectives. This limit exists even for large models like Tongyi-DeepResearch. Specifically, MemAct-RL shows strong generalization to unseen task complexities. Although it was trained on tasks with at most three objectives, it remains effective for up to eight objectives. It achieves 54.3\% accuracy in the 8-objective setting, which is significantly surpassing the 39.3\% achieved by Search-R1.

\paragraph{Domain Transfer Performance.}
MemAct consistently remains stable on simpler tasks like 2Wiki where basic reasoning is enough without any explicit memory action. The advantage of the model becomes much clearer as the reasoning complexity increases, as shown in Table~\ref{tab:main_results}. Additionally, the performance on BrowseComp-plus shows that MemAct generalizes well to new tool environments even when the underlying web corpus is unfamiliar.

\begin{table}[t]
\centering
\setlength{\tabcolsep}{2.6pt} 
\renewcommand{\arraystretch}{0.9}

\resizebox{0.99\linewidth}{!}{
\begin{tabular}{l|cc|cc|cc}
\toprule
 & \multicolumn{6}{c}{\textbf{Number of Objectives}} \\
\cmidrule(lr){2-7}
\textbf{Model}
 & \multicolumn{2}{c|}{\textbf{2-obj.}}
 & \multicolumn{2}{c|}{\textbf{4-obj.}}
 & \multicolumn{2}{c}{\textbf{8-obj.}} \\
\cmidrule(lr){2-3} \cmidrule(lr){4-5} \cmidrule(lr){6-7}
 & \textbf{Task} & \textbf{Mem.}
 & \textbf{Task} & \textbf{Mem.}
 & \textbf{Task} & \textbf{Mem.} \\
\midrule

\multicolumn{7}{l}{\textit{Ref Models}} \\
Qwen3-235B     
 & 5.3  & 0.0
 & 9.6  & 0.0
 & 18.2 & 0.0 \\

Tongyi-30B     
 & 5.4  & 0.0
 & 14.2 & 0.0
 & 33.2 & 0.0 \\

\midrule
\multicolumn{7}{l}{\textit{7B Models}} \\
Search-R1      
 & 7.7  & 0.0
 & 16.5 & 0.0
 & 23.5 & 0.0 \\

MemAct-SFT     
 & 12.0 & 1.6
 & 18.7 & 2.4
 & 21.8 & 2.8 \\

\textbf{MemAct-RL}
 & 13.6 & 2.1
 & 20.6 & 3.3
 & 25.2 & 3.7 \\

\midrule
\multicolumn{7}{l}{\textit{14B Models}} \\
Search-R1      
 & 7.5  & 0.0
 & 14.4 & 0.0
 & 21.3 & 0.0 \\

MemAct-SFT     
 & 10.4 & 1.5
 & 15.3 & 2.2
 & 22.1 & 3.7 \\

\textbf{MemAct-RL}
 & 8.7  & 1.4
 & 14.6 & 2.1
 & 20.2 & 3.9 \\

\bottomrule
\end{tabular}
}

\caption{\textbf{Tool usage statistics.} ``Task'' denotes task-related tool calls (e.g., search); ``Mem.'' denotes memory management actions (\texttt{Prune\&Write}).}
\label{tab:tool_usage_profile}

\end{table}

\begin{figure}[t!]
  \centering
  \includegraphics[width=0.995\linewidth,clip,trim={6bp 0 0 0}]{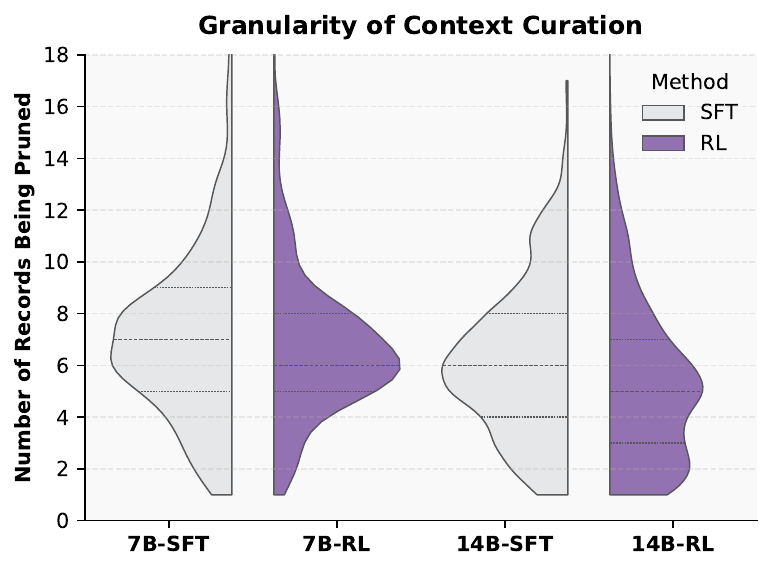}

\caption{
    \textbf{Pruning granularity distribution.} 
    Each violin shows the distribution of records pruned per \texttt{Prune\&Write} action ($|\mathcal{I}_{\text{target}}|$) on multi-objective tasks. 
    RL-trained policies exhibit lower variance than SFT, indicating convergence toward consistent strategies. The 14B-RL model shows a bimodal pattern with peaks at fine-grained ($\sim$2) and coarse-grained ($\sim$6) pruning.
}
\label{fig:behavior_analysis}
\end{figure}

\paragraph{Model-Specific Memory Strategies.}
MemAct automatically discovers strategies tailored to the capacity of each base model, as shown in Table~\ref{tab:tool_usage_profile} and Fig.~\ref{fig:behavior_analysis}, with further examples provided in Table~\ref{tab:memory_action_examples}.
\begin{itemize}[leftmargin=*, noitemsep, topsep=2pt]
    \item \textbf{7B Model:} For the 7B model, RL training leads to more frequent memory actions to handle its limited context capacity. In challenging 8-objective tasks, the action frequency increases notably from 2.8 to 3.7. Fig.~\ref{fig:behavior_analysis} shows that this model follows a consistent strategy by removing about 6 records per action to maintain stability.
    \item \textbf{14B Model:} The 14B model learns a strategy that separates ongoing research from task completion. As shown by the bimodal distribution in Fig.~\ref{fig:behavior_analysis}, this model performs fine-grained pruning (about 2 records) to remove irrelevant context during reasoning. In contrast, it performs coarse-grained pruning (about 6 records) to clear intermediate steps once a sub-objective is finished. This approach balances the need for detailed information with the goal of saving context space.
\end{itemize}

\section{Conclusion}
This paper presents Memory-as-Action (MemAct), a framework that internalizes context curation as a learnable capability by treating working memory management as explicit policy actions. To reconcile dynamic context updates with reinforcement learning, we introduced Dynamic Context Policy Optimization (DCPO), which ensures logical consistency by restructuring trajectories into independent segments at memory edit points. Our empirical results demonstrate that MemAct-RL-14B establishes a superior Pareto frontier for accuracy and efficiency, matching the performance of models over 16$\times$ larger while reducing average context length by 51\% and significantly improving end-to-end inference latency. Crucially, our analysis reveals that models autonomously discover specialized, capacity-aware strategies, adapting their memory action intensity to maintain a focused reasoning trace. Taken together, our results demonstrate that autonomous context curation can be internalized as a learnable skill, providing a fundamental and scalable architectural building block for agentic behavior in long-horizon reasoning processes.

\section{Limitations} \label{ssec:limitations}

While MemAct shows that context management can be learned, the current approach still faces challenges common in reinforcement learning for agents. The framework relies on sparse rewards from the final output, which makes it difficult to accurately assign credit to specific memory actions. In tasks that require long-horizon reasoning, the model might accidentally delete information that only becomes relevant later in the process. However, our analysis shows an intrinsic coupling between memory behavior and reasoning. This connection suggests that the MemAct paradigm could eventually help solve credit assignment issues in agent workflows. Furthermore, our current optimization method uses a random sampling algorithm that treats all memory operations as equally important. Since we do not yet use posterior methods to identify key steps, the process may allocate resources to segments with less information, which limits training efficiency in complex scenarios.

Regarding information fidelity, working memory management involves a trade off between context length and density. Because this compression process is lossy, the system cannot recover original data once details are summarized. This constraint defines the boundary and future potential of our study. Since local memory cannot maintain infinite precision, we see this approach as complementary to existing system-level infrastructures. Our priority is to verify the core mechanisms of memory actions within the standard context window to establish a principled interface at the decision layer. Future work can reduce the lossy nature of compression by expanding this action space, such as adding selective retrieval from external stores or tiered caching, to combine learned curation with scalable, high precision infrastructure.
\section{Acknowledgements}
This work is supported by the National Key R\&D Program of China (No. 2023YFC3310700) and the National Natural Science Foundation of China (No. 2576030). 

\bibliography{custom}

\appendix
\newpage
\clearpage
\section{Appendix}
\label{sec:appendix}

\subsection{Pseudocode for DCPO}
\label{sec:appendix_dcpo}
\begin{algorithm}[H]
\caption{DCPO Training Loop}
\label{app:DCPO}

\begin{algorithmic}[1]
\small
\STATE \textbf{Input:} Initial policy $\pi_\theta$, dataset $\mathcal{D}$, environment $\mathcal{E}$, \\
\hspace{2.2em} trajectories per prompt $N_{\text{traj}}$, segments per prompt $N_{\text{seg}}$
\STATE \textbf{Output:} Optimized policy $\pi_\theta$

\WHILE{not converged}
    \STATE Sample a batch of prompts $\mathcal{U} \sim \mathcal{D}$
    \STATE $\mathcal{B} \leftarrow \emptyset$ \quad \COMMENT{Global training batch}
    \STATE $A_{\text{map}} \leftarrow \{\}$ \quad \COMMENT{Map: trajectory $\to$ advantage}

    \FORALL{$u \in \mathcal{U}$}
        \STATE $\mathcal{T}_u \leftarrow \emptyset$
        \FOR{$n = 1$ \TO $N_{\text{traj}}$}
            \STATE Generate a trajectory $\tau$ from prompt $u$ and obtain reward $R(\tau)$
            \STATE $\mathcal{T}_u \leftarrow \mathcal{T}_u \cup \{(\tau, R(\tau))\}$
        \ENDFOR

        \STATE // \textit{Compute Advantage}
        \STATE $\mu_u \leftarrow \text{mean}(\{R \mid (\cdot, R) \in \mathcal{T}_u\})$
        \STATE $\sigma_u \leftarrow \text{std}(\{R \mid (\cdot, R) \in \mathcal{T}_u\}) + \epsilon$
        \FORALL{$(\tau, R) \in \mathcal{T}_u$}
            \STATE $A_{\text{map}}[\tau.\text{id}] \leftarrow (R - \mu_u) / \sigma_u$
        \ENDFOR

        \STATE $\Sigma_{u} \leftarrow \emptyset$ \quad \COMMENT{Segment pool for prompt $u$}
        \FORALL{$(\tau, \cdot) \in \mathcal{T}_u$}
            \STATE Identify memory edit points $\{t^{\text{mem}}_k\}$ in $\tau$
            \STATE Set $t^{\text{mem}}_0 \gets 0$, $t^{\text{mem}}_{K+1} \gets T$
            \FOR{$i = 0$ \TO $K$}
                \STATE $C_i \gets H_{t^{\text{mem}}_i}$ \quad \COMMENT{Context prefix}
                \STATE $\mathbf{y}_i \gets (y_t)_{t = t^{\text{mem}}_i + 1}^{t^{\text{mem}}_{i+1}}$ \quad \COMMENT{Segment generation}
                \STATE $\text{input\_ids} \gets \mathrm{tokenize}(C_i) \oplus \mathbf{y}_i$
                \STATE $m^{\sigma_i} \gets [0,\ldots,0,1,\ldots,1]$ \quad \COMMENT{Mask for $\mathbf{y}_i$, where $|0| = |\mathrm{tokenize}(C_i)|$}
                \STATE Append $(\text{input\_ids}, m^{\sigma_i}, \tau.\text{id})$ to $\Sigma_{u}$
            \ENDFOR
        \ENDFOR

        \STATE $\mathcal{B}_u \gets \text{Sample } N_{\text{seg}} \text{ segments from } \Sigma_{u}$
        \STATE $\mathcal{B} \leftarrow \mathcal{B} \cup \mathcal{B}_u$
    \ENDFOR

    \STATE $\mathcal{L}(\theta) \leftarrow \text{ComputePolicyLoss}(\mathcal{B}, A_{\text{map}}, \pi_\theta)$
    \STATE Update policy $\pi_\theta$ using $\nabla_\theta \mathcal{L}(\theta)$
\ENDWHILE

\STATE \textbf{return} $\pi_\theta$
\end{algorithmic}
\end{algorithm}

\subsection{SFT Cold-start Data Construction}

\label{app:sft_cold_start}

\paragraph{Staged Prompting for Cold-start Supervision.}
To initialize the memory policy, we construct a cold-start SFT dataset using a staged prompting protocol with DeepSeek-V3.1. The main motivation is that frontier models do not reliably invoke memory actions in long-horizon trajectories when left fully unconstrained. Typical failure modes include ignoring the memory tool, repeatedly invoking it without effective state updates, or losing coherence after a memory edit. To improve the success rate of valid memory trajectories, we inject different levels of intervention according to the current context length. When the accumulated context remains below a soft threshold $\tau_{\text{soft}}$, the model acts autonomously. When the context length falls between $\tau_{\text{soft}}$ and $\tau_{\text{hard}}$, we append a soft reminder encouraging the model to consider whether a memory update is needed. Once the context exceeds $\tau_{\text{hard}}$, we append a stronger instruction that explicitly prioritizes memory updating. After trajectory generation, we retain only successful trajectories whose final answers are correct, and remove the injected intervention messages from the final SFT data. In this way, the resulting dataset provides supervision for memory-aware behavior while avoiding dependence on explicit prompting cues at training time.

\begin{algorithm}[t]
\caption{Staged Prompting for Cold-start Data Construction}
\label{alg:sft_cold_start_simple}
\begin{algorithmic}[1]
\small
\STATE \textbf{Input:} task set $\mathcal{T}$, teacher model $M$, soft threshold $\tau_{\text{soft}}$, hard threshold $\tau_{\text{hard}}$
\STATE \textbf{Output:} SFT dataset $\mathcal{D}_{\text{sft}}$

\STATE Initialize $\mathcal{D}_{\text{sft}} \leftarrow \emptyset$

\FORALL{task $x \in \mathcal{T}$}
    \STATE Initialize an empty trajectory $\tau$
    \STATE Initialize the working context with the task instruction

    \FOR{$k = 1$ to $T_{\max}$}
        \STATE Measure the current context length $L$

        \IF{$L < \tau_{\text{soft}}$}
            \STATE Use the current context directly
        \ELSIF{$\tau_{\text{soft}} \leq L < \tau_{\text{hard}}$}
            \STATE Append a soft reminder to consider memory updating
        \ELSE
            \STATE Append a strong reminder to prioritize memory updating
        \ENDIF

        \STATE Let the teacher model generate the next action
        \STATE Execute the action and update the trajectory and working context

        \IF{the task terminates}
            \STATE \textbf{break}
        \ENDIF
    \ENDFOR

    \IF{the final answer is correct}
        \STATE Remove injected reminder messages from the recorded prompts
        \STATE Add the cleaned trajectory to $\mathcal{D}_{\text{sft}}$
    \ENDIF
\ENDFOR

\STATE \textbf{return} $\mathcal{D}_{\text{sft}}$
\end{algorithmic}
\end{algorithm}

\subsection{Datasets Statistics} \label{ssec:training_data}

Training data is generated from HotpotQA~\citep{yang2018hotpotqa} and Asearcher~\citep{gao2025turnsunlockinglonghorizonagentic} using the staged prompting protocol. The SFT phase uses 930 accurate examples, which are divided into 3,000 training segments to teach basic memory management rules. The RL phase scales to 10,240 trajectories to improve the policy. As shown in Table~\ref{tab:train_data_stat}, all training samples are limited to at most three objectives. This limit ensures that performance gains on tasks with 4 to 8 objectives reflect true generalization rather than just memorizing training patterns.

\begin{table}[ht]
\centering
\small
\setlength{\tabcolsep}{3.8pt}
\caption{\textbf{Training Dataset Composition.} Categorization of training instances by the number of reasoning objectives across SFT and RL phases.}
\label{tab:train_data_stat}
\setlength{\tabcolsep}{8pt}
\renewcommand{\arraystretch}{1.1}
\begin{tabular}{l|ccc|r}
\toprule
\textbf{Phase} & \textbf{1-Obj} & \textbf{2-Obj} & \textbf{3-Obj} & \textbf{Total} \\
\midrule
SFT   & 779   & 113   & 38    & 930   \\
RL    & 8,192 & 1,025 & 1,023 & 10,240 \\
\bottomrule
\end{tabular}
\end{table}

\subsection{Additional Results}

We provide additional experimental data to further evaluate the performance of smaller models and the specific behaviors of different model configurations.

\begin{table}[htbp]
\centering
\footnotesize 
\caption{\textbf{Performance comparison of 7B models on tasks with multiple objectives.} Accuracy represents the average success rate for individual sub objectives. Cost values indicate the consumption of input tokens ($\times 10^4$). Bold and underlined numbers show the best and second best results.}
\label{tab:appendix_7b}
\setlength{\tabcolsep}{1.2pt} 
\begin{tabular}{l ccccc c}
\toprule
\multirow{2}{*}{\textbf{Method}} & \multicolumn{5}{c}{\textbf{Multi Objective Tasks}} & \textbf{Cost} \\
\cmidrule(lr){2-6}
 & \textbf{2 obj.} & \textbf{4 obj.} & \textbf{6 obj.} & \textbf{8 obj.} & \textbf{Avg.} & \textbf{($\times 10^4$)} \\ 
\midrule
Qwen2.5 7B & 0.295 & 0.218 & 0.196 & 0.198 & 0.227 & \textbf{1.64} \\
Search R1 7B & \underline{0.553} & 0.409 & 0.383 & 0.341 & 0.422 & 11.97 \\
MemAct SFT 7B & 0.535 & \underline{0.431} & \underline{0.404} & \underline{0.366} & \underline{0.434} & \underline{10.81} \\
MemAct RL 7B & \textbf{0.598} & \textbf{0.493} & \textbf{0.434} & \textbf{0.416} & \textbf{0.485} & 11.97 \\ 
\bottomrule
\end{tabular}
\end{table}

\begin{table}[htbp]
\centering
\footnotesize
\caption{\textbf{Action statistics for MemAct across varying levels of task complexity.} Chain of Thought and Mem. represent the average token lengths for the reasoning prior to memory tools and the memory actions, respectively. Pruned Actions shows the average number of past actions removed from the context during each memory update. Numbers in parentheses indicate the change in RL compared to the SFT baseline.}
\label{tab:action_stats_compact}
\setlength{\tabcolsep}{1.1pt} 
\renewcommand{\arraystretch}{1.}
\begin{tabular}{l c ccc}
\toprule
\textbf{Method} & \textbf{Obj.} & \textbf{CoT Len.} & \textbf{Mem. Len.} & \textbf{Pruned Actions} \\
\midrule
\multirow{3}{*}{7B-SFT}  & 2 obj & 27.2 & 156.8 & 6.2 \\
                         & 4 obj & 17.0 & 165.6 & 6.9 \\
                         & 8 obj & 17.0 & 180.8 & 7.1 \\
\cmidrule(lr){1-5}
\multirow{3}{*}{7B-RL}   & 2 obj & 28.1 (+0.9) & 171.1 (+14.3) & 6.2 (0.0) \\
                         & 4 obj & 20.6 (+3.6) & 176.3 (+10.7) & 6.5 (-0.4) \\
                         & 8 obj & 21.8 (+4.8) & 198.7 (+17.9) & 7.0 (-0.1) \\
\midrule
\multirow{3}{*}{14B-SFT} & 2 obj & 29.4 & 167.1 & 5.9 \\
                         & 4 obj & 32.7 & 161.5 & 6.1 \\
                         & 8 obj & 31.0 & 186.5 & 5.8 \\
\cmidrule(lr){1-5}
\multirow{3}{*}{14B-RL}  & 2 obj & 35.2 (+5.8) & 159.7 (-7.4) & 5.6 (-0.3) \\
                         & 4 obj & 43.4 (+10.7) & 162.3 (+0.8) & 5.6 (-0.5) \\
                         & 8 obj & 35.7 (+4.7) & 191.1 (+4.6) & 5.2 (-0.6) \\
\bottomrule
\end{tabular}
\end{table}

\paragraph{Results for 7B Model Variants.}
The results in Table~\ref{tab:appendix_7b} show that MemAct RL 7B maintains a clear advantage over the baseline models in all settings. Specifically, MemAct RL 7B achieves the highest average accuracy of 0.485, which confirms that the memory action framework remains effective even on smaller language models. While the original Qwen2.5 7B model shows the lowest token consumption, this is primarily because it lacks long range reasoning capabilities. The base model fails to sustain the reasoning process required for complex tasks and terminates early. This behavior leads to both lower success rates and reduced token usage.
\paragraph{Analysis of Memory Strategies.}
The results in Table~\ref{tab:action_stats_compact} reveal distinct strategies for managing internal context between 7B and 14B models. Here, the Chain of Thought length specifically refers to the average length of the reasoning sequence generated immediately before each memory tool call. The data indicates that 14B models generally produce longer Chain of Thought sequences than 7B models, with reinforcement learning significantly increasing this reasoning depth. In contrast, 7B models maintain shorter reasoning lengths but rely more heavily on explicit memory storage. This is evidenced by the RL 7B model producing the longest memory records at 198.7 tokens for 8 objective tasks. Furthermore, the higher value in the Pruned Actions column shows that 7B models remove a greater number of past actions in a single memory update. This pattern suggests that smaller models perform more aggressive context clearing, likely because they need to purge more historical information at once to maintain focus given their limited processing capacity.

\paragraph{Additional Evaluation with Rule-based F1.}
We further report rule-based F1 scores as a complementary evaluation metric on the multi-objective QA benchmark. As shown in Table~\ref{tab:metric_compare}, the relative ranking of methods remains consistent under both evaluation protocols. Although the absolute F1 values are in general slightly lower, likely because model responses are often expressed as complete sentences rather than short phrase-level spans, MemAct-14B consistently outperforms Search-r1-14B across all difficulty levels under both LLM-Judge and rule-based F1. This result suggests that the observed advantage of MemAct is robust to the choice of evaluation metric.

\begin{table}[t]
\centering
\small
\setlength{\tabcolsep}{3pt}
\caption{LLM-Judge and rule-based F1 on multi-objective QA.}
\label{tab:metric_compare}
\resizebox{\columnwidth}{!}{
\begin{tabular}{llcccc}
\toprule
Method & Metric & 2-obj. & 4-obj. & 6-obj. & 8-obj. \\
\midrule
Search-r1-14B & LLM-Judge    & 0.625 & 0.546 & 0.493 & 0.393 \\
Search-r1-14B & Rule-based F1 & 0.558 & 0.524 & 0.509 & 0.428 \\
MemAct-14B    & LLM-Judge    & 0.660 & 0.591 & 0.570 & 0.543 \\
MemAct-14B    & Rule-based F1 & 0.572 & 0.541 & 0.540 & 0.522 \\
\bottomrule
\end{tabular}
}
\end{table}
\paragraph{PPO-based Variant of DCPO.}
We further examine whether DCPO generalizes beyond its default reinforcement learning backbone by implementing a PPO-based variant. In this variant, the underlying policy optimization algorithm is replaced with PPO, while the same context-segmentation mechanism is retained for handling dynamically edited contexts. As reported in Table~\ref{tab:ppo_variant_dcpo}, the PPO-based variant achieves competitive performance across all difficulty levels. Although its average result is slightly below that of the primary DCPO implementation, it remains consistently stronger than the Search-r1-14B baseline. This observation indicates that the effectiveness of DCPO does not depend on a specific policy optimization algorithm, but is mainly associated with its treatment of dynamic context restructuring during training.

\begin{table}[t]
\centering
\small
\setlength{\tabcolsep}{3pt}
\caption{Performance of a PPO-variant of DCPO.}
\label{tab:ppo_variant_dcpo}
\begin{tabular}{llccccc}
\toprule
Method & RL Method & 2-obj. & 4-obj. & 6-obj. & 8-obj. \\
\midrule
Search-r1-14B & GRPO & 0.625 & 0.546 & 0.493 & 0.393 \\
MemAct-14B & DCPO & 0.660 & 0.591 & 0.570 & 0.543 \\
MemAct-14B & DCPO-PPO  & 0.655 & 0.583 & 0.561 & 0.551 \\
\bottomrule
\end{tabular}
\end{table}
\paragraph{Additional Results on Qwen3-4B.}
We further evaluate MemAct on a more recent open-weight backbone, Qwen3-4B-2507-Instruct, to examine whether the proposed framework transfers beyond the main Qwen2.5-based setting. As shown in Table~\ref{tab:qwen3_4b_results}, MemAct improves the base Qwen3-4B model after both supervised initialization and reinforcement learning.

\begin{table}[t]
\centering
\small
\setlength{\tabcolsep}{4pt}
\caption{Additional results on Qwen3-4B on the multi-objective QA benchmark.}
\label{tab:qwen3_4b_results}
\begin{tabular}{lccccc}
\toprule
Method & 2-obj. & 4-obj. & 6-obj. & 8-obj. & Avg. \\
\midrule
MemAct-4B-SFT    & 0.642 & 0.560 & 0.485 & 0.380 & 0.517 \\
MemAct-4B-RL & 0.694 & 0.608 & 0.525 & 0.431 & 0.565 \\
\bottomrule
\end{tabular}
\end{table}
\paragraph{KV-Cache and Latency Breakdown.}
The existing results provide a plausible systems-level account of the observed efficiency gains. First, MemAct substantially reduces the active context processed at each step. In the main results, MemAct-RL-14B operates with an average active context of only 3{,}500 tokens per step and a total token cost of $8.2 \times 10^4$, compared with $19.3 \times 10^4$ for Search-R1-14B and $16.7 \times 10^4$ for Qwen3-235B. This reduction directly lowers prefill overhead and also alleviates the repeated decoding cost induced by overgrown histories. Second, memory editing remains sparse relative to the overall interaction trajectory, which limits the frequency of prefix disruption. For example, on 8-objective tasks, MemAct-RL-7B executes 25.2 task-related tool calls but only 3.7 memory actions on average, while MemAct-RL-14B executes 20.2 task-related tool calls with 3.9 memory actions. As a result, most generation phases still proceed over locally stable prefixes, so the practical benefit of cache reuse can largely be preserved despite occasional in-place edits. Taken together, these results suggest that the latency advantage of MemAct mainly arises from reducing the average prefill burden while preserving sufficiently stable decoding prefixes between sparse memory updates.

\subsection{Tool Scheme and Prompt Template}
\label{app:tool_prompt}
This section provides the complete tool schemas and system prompt templates used during training and inference.

\subsubsection{Memory Management Tool Schema}
\label{app:tool_schema}

\begin{tcolorbox}[
    breakable,
    colback=gray!5,
    colframe=gray!70,
    boxrule=0.5pt,
    arc=1pt,
    left=2pt, right=2pt, top=2pt, bottom=2pt,
    fontupper=\footnotesize,
    before skip=4pt,
    after skip=4pt
]
\textbf{Function:} \texttt{prune\_context}\\[2pt]
\textbf{Description:} Manages conversation history by removing redundant or superseded tool calls while preserving essential information. Use this tool to maintain context efficiency when the conversation becomes too long or contains outdated information.\\[2pt]
\textit{When to use:} Context exceeds token limits, information is superseded by newer results, or completed investigations need archival.\\[2pt]
\textit{Process:} First summarize key outcomes in \texttt{memory}, then specify which tool call IDs to remove via \texttt{delete\_ids}.\\[4pt]

\textbf{Parameters:}
\begin{itemize}[leftmargin=10pt, noitemsep, topsep=0pt]
    \item \texttt{memory} (string, required): 
    Detailed summary of key facts and current status from the tool calls being removed. It must contain sufficient information to continue reasoning.
    \item \texttt{delete\_ids} (array[string], required): 
    List of tool call IDs to remove from conversation history. These should be redundant or superseded tool calls.
\end{itemize}
\end{tcolorbox}

\subsubsection{System Instruction Template}
\label{app:system_prompt}
The following instruction template is prepended to every task query during both supervised fine-tuning and reinforcement learning:

\begin{tcolorbox}[
    breakable,
    colback=white,
    colframe=black!60,
    boxrule=0.5pt,
    arc=1pt,
    left=3pt, right=3pt, top=3pt, bottom=3pt,
    fontupper=\footnotesize,
    before skip=4pt,
    after skip=4pt
]
You are an Information Seeking Master. Your task is to thoroughly seek the internet for information and provide accurate answers to questions. You will not give up until you find the corresponding information.\\[4pt]

As you proceed, adhere to the following principles:\\[2pt]
\textbf{1. Persistent Actions for Answers}: You will engage in many interactions, delving deeply into the topic to explore all possible aspects.\\[2pt]
\textbf{2. Repeated Verification}: Before presenting a Final Answer, you will cross-check the information you have gathered to confirm its reliability.\\[2pt]
\textbf{3. Attention to Detail}: You will carefully analyze each information source to ensure that all data is current and relevant.\\[4pt]

\textbf{Core Tool: prune\_context}\\[2pt]
\begin{itemize}[leftmargin=10pt, noitemsep, topsep=0pt]
    \item You have access to \texttt{prune\_context} for active context management. Its function is to summarize specific tool calls into memory paragraphs and delete them after execution.
    \item Treat this as a core reasoning tool. Use it throughout every step of your reasoning process to maintain focus and clarity.
\end{itemize}
\vspace{2pt}

\textbf{Context Management Policy}\\[2pt]
\begin{itemize}[leftmargin=10pt, noitemsep, topsep=0pt]
    \item Keep the active context logically consistent.
    \item Continuously summarize outcomes and prune superseded steps to maintain a high signal-to-noise ratio.
\end{itemize}
\vspace{2pt}

\textbf{Memory Structure:}\\[2pt]
\begin{itemize}[leftmargin=10pt, noitemsep, topsep=0pt]
    \item \texttt{[OBJECTIVE]} The ultimate goal of the user.
    \item \texttt{[CONCLUSIONS/FACTS]} Verifiably true information gathered.
    \item \texttt{[STATUS]} The specific and actionable next step.
    \item \texttt{[ASSUMPTIONS]} Optional unverified premises used to proceed.
\end{itemize}
\end{tcolorbox}

\subsection{Case Study}
\label{app:case_study_section}
In this section, we provide detailed examples of memory actions. Table~\ref{tab:memory_action_examples} illustrates how the model uses the \texttt{Prune\&Write} tool to manage context during different reasoning stages. Table~\ref{tab:failure_cases} analyzes common failure modes where the model struggles with evidence ambiguity or memory hallucinations.

\begin{table*}[!htbp]

\footnotesize
\centering
\caption{\textbf{Examples of different memory curation strategies learned by MemAct.} These cases illustrate how the model adapts its pruning behavior to support final answer preparation, subtask transitions, and information consolidation across various reasoning stages.}
\label{tab:memory_action_examples}
\renewcommand{\arraystretch}{1.2}
\label{app:memory_cases}

\begin{tabularx}{\textwidth}{
    >{\raggedright\arraybackslash}p{2.5cm}
    >{\raggedright\arraybackslash}X
    >{\raggedright\arraybackslash}p{2.cm}
    >{\raggedright\arraybackslash}p{4.0cm}
    }
\toprule
\textbf{Chain-of-thoughts} & \textbf{Memory Content} & \textbf{Removed Ids}  & \textbf{Mechanism Analysis} \\
\midrule

Let me prune the context to maintain efficiency and summarize what I've learned about Groundation's connection to Jamaica.
&
\textbf{[OBJECTIVE]} Determine what Groundation and Jamaica have in common \newline
\textbf{[CONCLUSIONS/FACTS]} Groundation is a California-based reggae band... name comes from Rastafarian "groundings"... references Grounation Day... which commemorates Haile Selassie's... visit to Jamaica... The band plays roots reggae music... \newline
\textbf{[STATUS]} Ready to provide final answer... \newline
\textbf{[ASSUMPTIONS]} Groundation is a reggae band that draws heavily from Jamaican musical and cultural traditions
&
\texttt{4sNKCfUv}, \texttt{nNUehhvj}, ... \newline \textit{(Total 10)}
&
\textbf{Final Answer Preparation}: The operation prunes intermediate search steps. The structured memory records the synthesized findings and preserves the established links between the band's name, musical style, and Jamaican culture. \\

\midrule

Let me prune the context to keep the conversation focused and continue searching for the answers.
&
\textbf{[OBJECTIVE]} Answer three specific questions... \newline
\textbf{[CONCLUSIONS/FACTS]} Found information about Bargain Booze... but need to verify if it was founded in 1981... \newline
\textbf{[STATUS]} Need to search for specific information about Bargain Booze's founding year... then move to questions 2 and 3 \newline
\textbf{[ASSUMPTIONS]} Bargain Booze may be the answer to Q1 but requires verification
&
\texttt{Jrr7YuRY}, \texttt{9EnqZAXu}, ... \newline \textit{(Total 9)}
&
\textbf{Phase Transition}: The operation clears the context after the identification subtask. The structured memory retains the candidate answer and explicitly lists the remaining verification requirements (date and store count). \\

\midrule

Let me prune the context to maintain efficiency and continue with my research on the composers.
&
\textbf{[OBJECTIVE]} Determine which composer had their first notable composition released earlier... \newline
\textbf{[CONCLUSIONS/FACTS]} Josquin des Prez was active... in the 1470s and his works were published as early as 1502. Palestrina's first major publication was in 1554. Josquin's motet "Ave Maria..." is... one of his early works... \newline
\textbf{[STATUS]} Need to verify specific dates of Josquin's earliest compositions...
&
\texttt{t4qeJxRu}, \texttt{trWYuaUY}, ... \newline \textit{(Total 5)}
&
\textbf{Information Consolidation}: The operation removes raw search results. The structured memory retains specific dates and publication details for both entities to support the subsequent temporal comparison. \\

\midrule

Let me prune the context to remove the less relevant search results and continue with a more focused search.
&
\textbf{[OBJECTIVE]} Find the name of the unseen announcer on TV's University Challenge \newline
\textbf{[CONCLUSIONS/FACTS]} Initial searches have not yet identified the specific unseen announcer. ...hosted by Bamber Gascoigne... and Jeremy Paxman..., but these are the visible hosts... \newline
\textbf{[STATUS]} Need to find specific information about the voice announcer... focusing on the name Roger Tilling... \newline
\textbf{[ASSUMPTIONS]} The unseen announcer is likely a different person from the visible host...
&
\texttt{sxDSsMtk}, \texttt{ekbxCapH}, ... \newline \textit{(Total 8)}
&
\textbf{Research Refocusing}: The operation eliminates unproductive search paths regarding visible hosts. The updated memory records the negative results and formalizes the specific hypothesis (Roger Tilling) for the next search phase. \\

\midrule

Let me prune the context to maintain efficiency while continuing my research.
&
\textbf{[OBJECTIVE]} Answer two questions: Q1 about... Curuá River... and Silver River...; Q2 about legal tradition... where Sai Van Bridge is located \newline
\textbf{[CONCLUSIONS/FACTS]} Curuá River is in Brazil... Silver River appears to be in Michigan... Sai Van Bridge is in Macau... Macau was Portuguese territory... \newline
\textbf{[STATUS]} Need to verify Silver River location... and research Macau's legal system tradition \newline
\textbf{[ASSUMPTIONS]} Silver River likely exists in Michigan... Macau's legal system likely influenced by Portuguese tradition
&
\texttt{zX9vraas}, \texttt{X5Aouyfx}, ... \newline \textit{(Total 9)}
&
\textbf{Multi-Query Management}: The operation removes completed search segments. The structured memory separates findings by question number and distinguishes confirmed facts (locations) from unverified claims. \\

\bottomrule
\end{tabularx}
\end{table*}

\newpage
\clearpage

\begin{table*}[!htbp]
\footnotesize
\centering
\caption{\textbf{Analysis of representative failure cases during autonomous context management.} These examples show how unresolved ambiguity or missing evidence can lead to incorrect assumptions being stored in the memory and affecting the final reasoning outcome.}
\label{tab:failure_cases}
\renewcommand{\arraystretch}{1.2}

\begin{tabularx}{\textwidth}{
    >{\raggedright\arraybackslash}p{2.5cm}
    >{\raggedright\arraybackslash}X
    >{\raggedright\arraybackslash}p{2.cm}
    >{\raggedright\arraybackslash}p{4.0cm}
}
\toprule
\textbf{Chain-of-thoughts} & \textbf{Memory Content} & \textbf{Removed Ids} & \textbf{Mechanism Analysis} \\
\midrule

Let me prune the context to reduce noise. The author name “The Thing of It Is…” appears in two different search results, but most likely they refer to the same person. I’ll assume the journalist wrote it and finalize.
&
\textbf{[OBJECTIVE]} Identify the nationality of the author of “The Thing of It Is…” and determine the Western TV series associated with Pat Hingle’s friend. \newline
\textbf{[CONCLUSIONS/FACTS]} Search results show both William Goldman (novelist) and John Callaway (journalist) have works titled similarly. \newline
\textbf{[STATUS]} Conclude that the autobiography by John Callaway is the referenced work; proceed to answer. \newline
\textbf{[ASSUMPTIONS]} “The Thing of It Is…” is more likely an autobiography than a novel.
&
\texttt{aP9xLmQ2}, \texttt{vT3bKp11}, … \newline \textit{(Total 6)}
&
\textbf{Ambiguity Ignored}:  
The model encounters two different works with the same title but doesn’t resolve the conflict. Instead of checking further, it collapses the ambiguity and assumes both refer to the same author. This mistaken assumption gets written into memory and steers the rest of the reasoning in the wrong direction.\\
\midrule

Let me prune the context to maintain efficiency and consolidate my findings about who recorded “Something There.”
All searches so far point to the original film cast, but they do not surface any American actress born in 1956. 
&
\textbf{[OBJECTIVE]} Identify the American actress/singer born in 1956 who recorded “Something There.”\newline
\textbf{[CONCLUSIONS/FACTS]} Multiple searches confirm the original vocalists were Paige O’Hara (born 1954) and Robby Benson (born 1955). No evidence supports any 1956-born recording artist for this song.\newline
\textbf{[STATUS]} Conclude that Toni Braxton, notable cast member of Beauty and the Beast, must be the intended answer.\newline
\textbf{[ASSUMPTIONS]} Toni Braxton participated in soundtrack-related recordings and may have covered “Something There.”
&
\texttt{mxTV6UnX}, \texttt{TG6tXj9F}, \texttt{XbyiVBRr}, … \newline\textit{(Total 7)}
&
\textbf{Hallucination in Memory}:  
After repeated searches fail to produce any relevant result, the model feels compelled to fill the gap. It invents a connection that isn’t supported by any evidence and treats the guess as fact. Once this fabricated link is stored in memory, it drives the model toward a confident but entirely incorrect final answer.\\
\bottomrule
\end{tabularx}
\end{table*}

\end{document}